\documentclass[journal]{IEEEtran}
\usepackage{amssymb}
\usepackage{graphicx}
\usepackage{color}
\usepackage{amsmath}
\usepackage{multirow}
\usepackage{makecell}
\usepackage{bbding}
\usepackage{amsmath}
\usepackage{amsthm}
\ifCLASSINFOpdf
\else
\fi
\begin{document}
%
\title{DiffVein: A Unified Diffusion Network for Finger Vein Segmentation and Authentication}
%
%
%

\author{Yanjun~Liu, Wenming~Yang$^{*}$,~\IEEEmembership{Senior Member,~IEEE,} Qingmin~Liao,~\IEEEmembership{Senior Member,~IEEE,}
\thanks{Yanjun Liu, Wenming Yang and Qingmin Liao are with the Shenzhen International Graduate School, Tsinghua University, Shenzhen 518055,China.}
\thanks{Manuscript received XXX; revised XXX. 
Corresponding author: Wenming Yang (email: yang.wenming@sz.tsinghua.edu.cn).}}

%
%

\markboth{Journal of \LaTeX\ Class Files,~Vol.~14, No.~8, August~2015}%
{Shell \MakeLowercase{\textit{et al.}}: Bare Demo of IEEEtran.cls for IEEE Journals}
%



\maketitle

\begin{abstract}
Finger vein authentication, recognized for its high security and specificity, has become a focal point in biometric research. Traditional methods predominantly concentrate on vein feature extraction for discriminative modeling, with a limited exploration of generative approaches. 
Suffering from verification failure, existing methods often fail to obtain authentic vein patterns by segmentation.
To fill this gap, we introduce DiffVein, a unified diffusion model-based framework which simultaneously addresses vein segmentation and authentication tasks. 
DiffVein is composed of two dedicated branches: one for segmentation and the other for denoising. For better feature interaction between these two branches, we introduce two specialized modules to improve their collective performance.
The first, a mask condition module, incorporates the semantic information of vein patterns from the segmentation branch into the denoising process.
Additionally, we also propose a Semantic Difference Transformer (SD-Former), which employs Fourier-space self-attention and cross-attention modules to extract category embedding before feeding it to the segmentation task. In this way, our framework allows for a dynamic interplay between diffusion and segmentation embeddings, thus vein segmentation and authentication tasks can inform and enhance each other in the joint training. To further optimize our model, we introduce a Fourier-space Structural Similarity (FourierSIM) loss function, which is tailored to improve the denoising network's learning efficacy. Extensive experiments on the USM and THU-MVFV3V datasets substantiates DiffVein's superior performance, setting new benchmarks in both vein segmentation and authentication tasks.
\end{abstract}


%
\IEEEpeerreviewmaketitle

\section{Introduction}
With the proliferation of intelligent devices in both industrial and everyday settings, biometric recognition technologies are increasingly replacing conventional authentication methods such as passwords and keys, due to their superior security and convenience. Among these technologies, finger vein authentication emerges as a particularly promising method. This technique, which relies on the unique characteristics of human finger veins, offers several technical advantages over other biometric approaches. Notably, finger veins are located deep within the finger and their patterns are determined mostly by genetic factors, which makes them nearly impossible to replicate. Moreover, finger vein authentication presents a reduced risk of privacy violation compared to more prevalent methods like facial or fingerprint recognition. Consequently, finger vein-based authentication have drawn favourable attention from both the research community and commercial sectors alike.

The common practice in finger vein authentication involves capturing images of the finger's ventral side to analyze the vein patterns. This process relies on the principle that hemoglobin in the veins absorbs near-infrared light at specific wavelengths \cite{mancini1994validation}, which makes the veins stand out as dark lines against the lighter background of other tissues. Ideal image capture results in a clear and uninterrupted vein network, which are essential for accurate recognition. However, the quality of vein images can be compromised by factors such as inconsistent blood flow, varying vein depth, and potential occlusion by surrounding tissues. These factors can lead to images with veins that are either too bright or too faint, making them difficult to differentiate from adjacent background tissue. The overall performance of finger vein authentication systems, therefore, relies on consistently discerning and analyzing these vein patterns amidst the natural variability of imaging conditions. To address these challenges, sophisticated algorithms are essential for the precise extraction of vein patterns.

Traditional feature extraction methods for finger vein recognition typically focus on extracting either explicit or implicit features. Explicit features are often engineered based on human priors, which encompass information such as image intensity and contrast to distinguish vein regions from the background. The primary limitation of this approach is its suboptimal performance in extracting veins from poorly-captured finger images, leading to potential errors in regions where veins are not pronounced. Implicit features, on the other hand, involve encoding hand-crafted features or local vein image patches into matrices or vectors. These feature representations capture more discriminative features in high-dimensional spaces, thereby improving the separability of vein patterns. Recent trends in implicit feature-based recognition have turned to learning-based methods. With the powerful learning capabilities of deep networks, these approaches can obtain more effective and robust feature representations than traditional counterparts. Another advantage of deep learning-based methods is their ability to integrate feature extraction and classification into a unified end-to-end framework.

\begin{figure}[htbp]
\centering
\begin{minipage}{0.16\linewidth}\centering
\includegraphics[width=\linewidth]{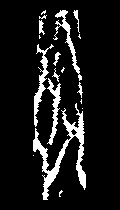}
\centerline{(a) RLT}
\end{minipage}
\hspace{10pt}
\begin{minipage}{0.16\linewidth}\centering
\includegraphics[width=\linewidth]{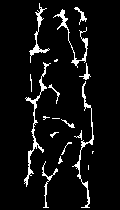}
\centerline{(b) LMC}
\end{minipage}
\hspace{10pt}
\begin{minipage}{0.16\linewidth}\centering
\includegraphics[width=\linewidth]{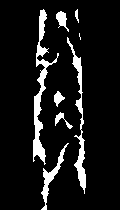}
\centerline{(c) WLD}
\end{minipage}
\hspace{10pt}
\begin{minipage}{0.16\linewidth}\centering
\includegraphics[width=\linewidth]{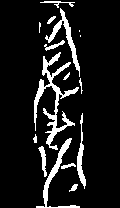}
\centerline{(d) Gabor filter}
\end{minipage}
\caption{Examples of finger vein segmentation based on traditional methods.}
\label{chap1:fig:dspseg}
\end{figure}

Despite these advancements, there has been limited research on the extraction of actual vein patterns, which are rich in semantic information on the geometric and topological characteristics of the finger veins. While some explicit feature extraction methods can generate vein networks by identifying specific patterns, such as valley-like \cite{miura2004feature, miura2007extraction} or line-like features \cite{kumar2011human, huang2010finger}, these local features often have more statistical than geometric relevance. This can lead to visual effects that diverge significantly from actual vascular structures, as shown in Fig. \ref{chap1:fig:dspseg}. 
While binary masks derived from precise vein segmentation may not necessarily enhance the finger vein recognition process as effectively as established explicit, implicit, and especially learning-based methods, they still hold significant research value. 
Given that finger veins are inherently tubular structures, research into their segmentation is particularly relevant for similar medical applications, such as the segmentation of blood vessels in CT and MRI images. Therefore, a key motivation of our work is to incorporate vein segmentation into the recognition process and to study a multi-task learning framework that integrates both tasks.

The Denoising Diffusion Probabilistic Model (DDPM) \cite{ho2020denoising} has gained significant attention for its capacity to produce high-quality and diverse images, as evidenced by large-scale models such as DALLE2 \cite{ramesh2022hierarchical}, Imagen \cite{saharia2022photorealistic}, and Stable Diffusion \cite{rombach2022high}. Originally applied in generative tasks without groundtruth data, diffusion models have also proven exceptionally effective in supervised settings, exemplified by their applications in super-resolution \cite{saharia2022image} and medical image segmentation \cite{wu2022medsegdiff, wu2023medsegdiff}. There have been adaptations of diffusion models for discriminative purposes in image classification \cite{han2022card, yang2023diffmic}, yet the finger vein field lacks such precedents. It is important to note that FV-GAN \cite{yang2019fv} does leverage the generative capabilities of GANs for domain adaptation between vein image and latent vein pattern spaces. However, its latent representations are tailored more towards discrimination tasks rather than the extraction of actual vein masks. This paper aims to bridge this gap by introducing a diffusion model-based generative framework for finger vein recognition and segmentation. 

Based on the motivations mentioned above, we propose a unified, diffusion model-based framework for vein segmentation and authentication named DiffVein. Specifically, it comprises four main modules: a segmentation network, a diffusion process, a denoising network, and a Semantic Difference Former (SD-Former). The segmentation network is designed to extract vein patterns from finger vein images and provide feature embedding priors for the diffusion process. The diffusion process then generates noisy variables from these priors, and the denoising network aims to predict the noise signal. The SD-Former introduces an innovative feature interaction strategy, enabling the flow of different semantic information within this multi-task learning framework.
To summarize, this paper presents three contributions listed as follows:
\begin{itemize}
\item We propose a unified diffusion model-based framework which achieve simultaneous learning of finger vein segmentation and authentication task. In this framework, denoising network is conditioned by vein segmentation, while the diffusion model provide  feedback on generated feature representations to the segmentation task. 
\item We propose a Semantic Difference Former (SD-Former) to extract the hidden category embedding in the denoising network and introduce it to the segmentation network. The module is implemented by Fourier space-based self-attention and cross-attention mechanism for better semantics separation in frequency-domain.
\item We propose a novel training loss for the denoising network based on Fourier-space Structural Similarity (FourierSIM), which takes into account the discrepancy of amplitude and phase angle between the Gaussian noise and the predicted noise.
\end{itemize}
\section{Related Works}
\subsection{Finger Vein Recognition} 
Extraction of vein features is the central part of the finger vein recognition and critically influences the overall performance. Over the last decade, extensive research has been conducted to develop methods for extracting robust and distinctive vein features. These methods fall into two primary categories: hand-crafted features and deep learning.
\subsubsection{Hand-crafted features}
Methods based on hand-crafted features were dominant in the early stage of vein recognition study and the research focus was mainly on the design of local direction and intensity features. Specifically, vein patterns are characterized by dark lines within the image, which correspond to regions of lower intensity values. To capture the local direction and frequency features of these patterns, several studies have employed Gabor filters with various orientations and scales \cite{wang2019weber, zhang2019adaptive, kumar2022finger} as convolution kernels. Furthermore, algorithms like Local Directional Code (LDC) \cite{meng2012finger} and Polarized Depth-weighted Binary Direction Coding (PWBDC) \cite{yang2019hybrid} have been developed to encode the local gradient orientation for more distinguishing features in the image.

For vein pattern intensity analysis, Repeated Line Tracking (RLT) \cite{miura2004feature} iteratively follows those dark lines by assessing the intensity contrast between adjacent regions. Local Maximum Curvature (LMC) \cite{miura2007extraction} locates veins by evaluating the curvature along the cross-sectional grayscale profile. Building upon LMC, Syarif et al. \cite{syarif2017enhanced} introduced an Enhanced Maximum Curvature (EMC) approach that integrates the Hessian matrix and histogram of oriented gradients for improved pattern extraction. Anatomy Structure Analysis based Vein Extraction (ASAVE) \cite{yang2012finger} further refines vein extraction by utilizing curvature analysis based on the characteristic valley shapes of vein cross-sections. In addition, Local Binary Patterns (LBP) \cite{wang2012finger} and Wide Line Detector (WLD) \cite{huang2010finger} algorithms have been explored for rule-based encoding of grayscale values. These methods, particularly when combined with manual feature extraction, offer robust detection of linear vein features.


\subsubsection{Deep learning}
The advent of deep learning has ignited a surge of research in finger vein recognition, with numerous studies exploring novel network architectures tailored for this application. Pioneering work by Hong et al. \cite{hong2017convolutional} demonstrated the applicability of the pre-trained VGG network to vein feature extraction, marking a significant milestone for deep learning in this field. Subsequent innovations include the development of specialized networks like FV-Net \cite{hu2018fv}, autoencoders \cite{hou2019convolutional}, siamese networks \cite{tang2019finger,fang2023finger}, and other novel architectures. Additionally, Generative Adversarial Networks (GANs) have shown promise to generate optimal binary vein masks ever since FV-GAN \cite{yang2019fv}. This generative model also serves as a solution to constructing synthetic finger vein image datasets.

A notable advancement in the field is the application of attention mechanism-based networks. The Transformer model \cite{vaswani2017attention}, and its finger vein recognition-specific adaptation FVT \cite{huang2022fvt}, have marked a significant step forward. Rather than modifying the conventional Transformer architecture, MRFBCNN \cite{wang2021finger} employs bilinear pooling to integrate first-order features with subsequent higher-order features generated by the attention mechanism, thereby improving the representation of vein features. In parallel, some studies have focused on enhancing recognition accuracy through feature fusion methods similar to attention mechanism. For instance, EIFNet \cite{song2022eifnet} innovatively extracts both explicit and implicit features via two separate networks, and Noh et al. \cite{noh2020finger} investigated a fusion strategy that combines vein texture with shape features. 

In addition to innovations in network architectures and modules, significant strides have been made in the development of novel loss functions specifically designed for biometric recognition, including Arccosine center loss \cite{hou2021arcvein}, Fusion loss \cite{ou2021fusion}, and dynamic regularized Center loss \cite{zhao2020finger}. Such loss functions are instrumental in guiding the network to learn more discriminative features, thereby improving the separability between different classes (inter-class distinction) while maintaining a high degree of similarity within the same class (intra-class compactness). 

Despite the advancements mentioned above, the inherent complexities of vein images, such as quality variations and ambiguity in vein localization, still present ongoing challenges.

\subsection{Finger Vein Segmentation}
The variability in blood flow and tissue structure beneath skin often results in inconsistent vein appearances in images, posing significant challenges for accurate finger vein segmentation. Both traditional methods and contemporary methods have been proposed to extract vein patterns and generate vein masks based on different paradigms and we will review some of the representative methods in the following paragraph.

\begin{figure}[htbp]
\centering
\begin{minipage}{0.4\linewidth}\centering
\includegraphics[angle=90, width=\linewidth]{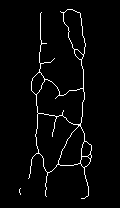}
\centerline{(a) ASAVE network}
\end{minipage}
\hspace{20pt}
\begin{minipage}{0.4\linewidth}\centering
\includegraphics[angle=90, width=\linewidth]{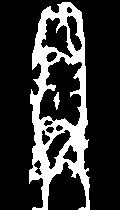}
\centerline{(b) PC}
\end{minipage}
\caption{Examples of thick and thin vein patterns extracted by traditional methods.}
\label{chap1:fig:thinthick}
\end{figure}

Tradition vein pattern extractors are mainly based on human-crafted features, which means they derive from recognition methods mentioned in the last section. These methods focused on extracting the geometric and structural patterns of finger veins such as directional features \cite{kumar2011human} and local curvature features \cite{miura2007extraction,syarif2017enhanced}, which inadvertently facilitated vein segmentation. However, these methods vary considerably in their ability to generate vein masks from identical images. For example, as illustrated in Fig. \ref{chap1:fig:thinthick}, the ASAVE network \cite{yang2017fingerv2} extracts thinner veins, while the vein patterns extracted by PC \cite{choi2009finger} are generally coarser and thicker. This discrepancy underscores the limitations of traditional approaches, which prioritize feature extraction over accurately representing vein patterns. In other words, the vein masks generated by these methods do not throw light on the actual position of embedded human veins. 

In the era of deep learning, the U-Net architecture \cite{cciccek20163d} has proven to be a powerful tool for medical image segmentation. Actually, it is still effective in modalities akin to the near-infrared images central to our study. Building on this influential framework, EIFNet \cite{song2022eifnet} introduced the first human-annotated finger vein segmentation dataset and implemented a U-Net-inspired mask generation module to capture vein patterns. Despite EIFNet's contributions, research specifically targeting finger vein segmentation remains sparse. This scarcity underlines the potential for drawing valuable insights from related works in the field of medical imaging, particularly studies focused on segmenting complex tubular structures, such as coronary arteries and retinal vessels. 

Kernel-based design and architecture-based design are the two main streams for such studies. Methods such as dilated \cite{yu2017dilated} and deformable \cite{dai2017deformable} convolutions, modify the shape of convolutional kernels to capture diverse local structural features, overcoming the limitations of classic CNNs' receptive field. DUNet \cite{jin2019dunet} and DCUNet \cite{yang2022dcu} incorporate deformable convolutions to dynamically adapt to the geometric variations in vascular structures. On the other hand, architecture-based designs focus on crafting special network architectures, which enhances the model's capability to discern and interpret the morphological features of tubular structures. PointScatter \cite{wang2022pointscatter} introduces point set representations for greater flexibility. 
TransUNet \cite{chen2021transunet} combines U-Net with Transformer architecture for better capacity of modeling global contexts while meriting from CNNs’ inherent ability of precise localization. 
To address this possible convergence difficulties due to unrestricted geometric learning, DSCNet \cite{qi2023dynamic} integrates prior knowledge of tubular structures by adopting a snake-like convolution operator to enhance the perception of tubular features.

\subsection{Diffusion Models}
Machine learning algorithms for solving classification and regression problems typically fall into two paradigms: discriminative and generative. Discriminative methods directly model the decision boundary of the task, while generative methods generally learn to model data distribution before transforming the target task into a problem of Maximum Likelihood Estimation (MLE). In recent years, generative models based on DDPM \cite{ho2020denoising} have attracted significant interest due to their training stability and outstanding capacity to generate high-dimensional data such as images \cite{saharia2022photorealistic, rombach2022high}, videos \cite{singer2023makeavideo}, and 3D objects \cite{lin2023magic3d}. These models initially transform the target data into a noisy domain by introducing Gaussian noise, and then recover the original data through a reverse diffusion process.

Early research focused on unconditional generative modeling \cite{dhariwal2021diffusion}, but the trend has shifted towards conditional generation that aligns images with specific semantics, such as labels \cite{yang2023diffmic} or text embeddings \cite{singer2023makeavideo, lin2023magic3d}. In the context of finger vein recognition, the capabilities of DDPM-like models are yet to be fully explored. Recent studies in classification \cite{han2022card, yang2023diffmic} suggest that diffusion models can effectively learn feature representations for reduced-dimensionality tasks, which is consistent with the goals of finger vein recognition. Building on these insights, our work aims to integrate the finger vein authentication task into a diffusion-based generative framework.

\section{Methodolody}

\subsection{Overview} \label{sec:overview}
\begin{figure*}[htbp]
\centering
\includegraphics[width=\linewidth]{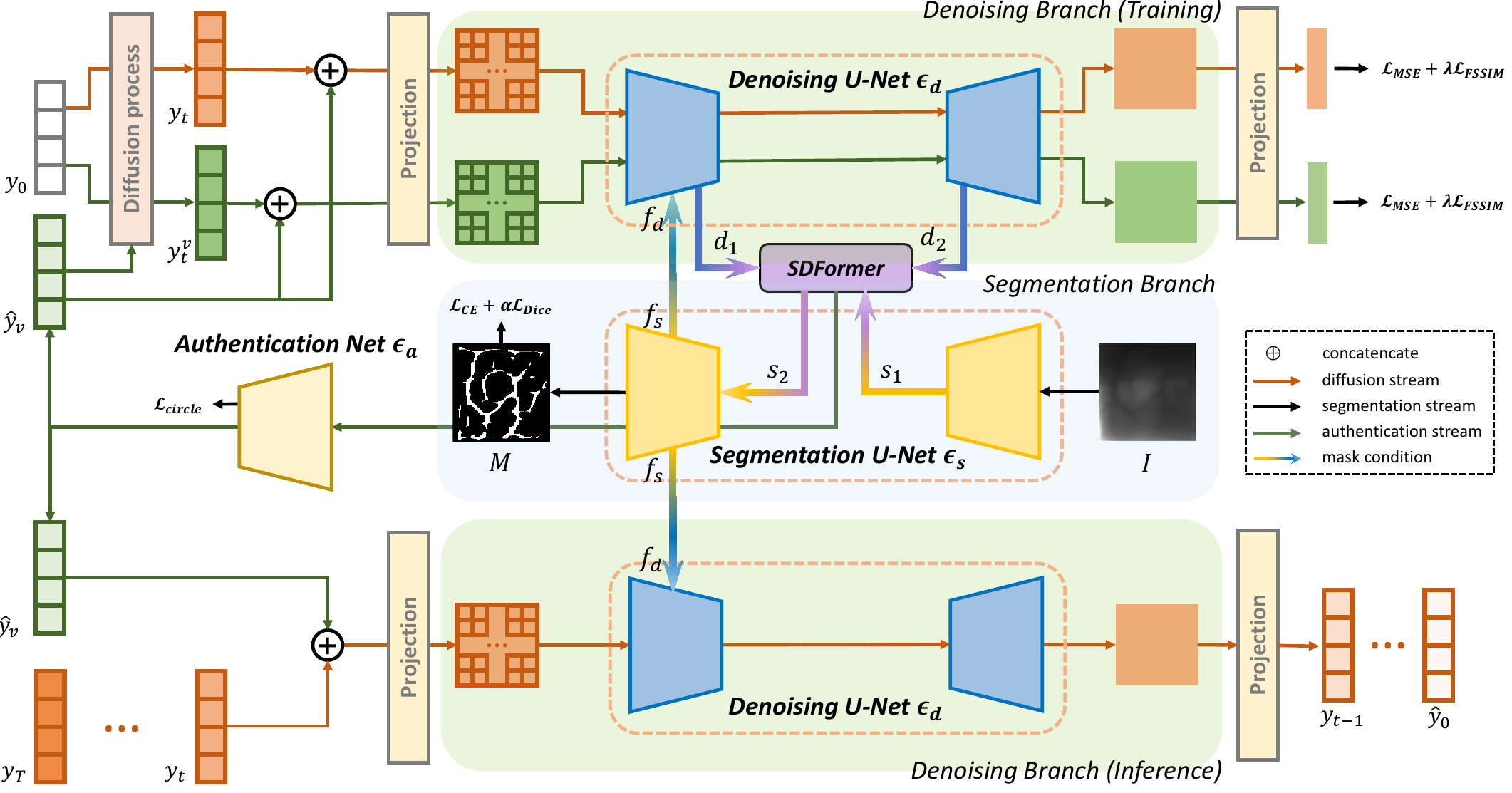}
\caption{The schematic illustration of DiffVein.}
\label{chap3:fig:overview}
\end{figure*}

Fig. \ref{chap3:fig:overview} shows the overall architecture of our proposed framework named DiffVein. It contains two U-Net-like networks for denoising and segmentation branches. Given an image of the human finger captured by NIR camera, the segmentation branch mainly aims to localize the vein regions in pixel level and segment them in a binary mask. Besides, this network also serves as the feature extractor to generate category priors for the authentication task. The details of this network are presented in Sec. \ref{chap3:sec:segment}.
The denoising branch is responsible for predicting the Gaussian noise added on the noisy inputs. Our diffusion model and denoising network is elaborated in Sec. \ref{chap3:sec:diffusion}. It should be noted that the upper and lower part of the framework illustrates the forward process for training and the reverse process for inference, respectively. In this figure, the noisier variables appear in darker colors than their counterparts with less noise. 

The segmentation feature maps from the expanding path in segmentation U-Net are supposed to provide an important guidance to the denoising network. Thus, we develop a mask condition module to introduce vein semantics to the denoising process, which is elaborated in Sec. \ref{chap3:sec:diffusion}. On the other hand, the segmentation task may potentially benefit from the implicit category information encoded in the denoising branch. Thus, the Semantic Difference Transformer (SD-Former) is proposed to extract semantic information from the denoising branch, and the details of this module are given in Sec.\ref{chap3:sec:sdformer}. To sum up, bidirectional information will flow across both branches in a complementary way which boosts the overall performance.
\subsection{Segmentation Branch} \label{chap3:sec:segment}
This branch primarily focuses on generating vein masks and producing conditional priors of finger's category for the diffusion process. Given an input finger image $I$, the segmentation network, denoted as $\epsilon_s$, is tasked with the prediction of the corresponding binary vein mask $M$. 
To optimize computational efficiency, $\epsilon_s$ is configured as a U-Net \cite{cciccek20163d} with modified input channels. The specific network architecture in Table \ref{chap3:tbl:seg_unet}.
Note that \textit{Convolution} in this table refers to a sequence of operations comprising convolution, batch normalization, and ReLU activation function. \textit{Convolution*} refers to a compound layer consisting of two consecutive \textit{Convolution} layers. In the expanding path, convolution layers with upsampling module are utilized to output the predicted vein mask. This segmentation task is supervised by binary cross entropy and Dice loss formulated as:
\begin{equation}
\mathcal{L}_{seg} = \mathcal{L}_{Dice} + \alpha \mathcal{L}_{CE}
\label{chap3:eq:seg_loss}
\end{equation}
where $\alpha$ is a weight factor. In this paper, $\alpha$ is set as 0.8.
\begin{table}[htbp]
\caption{Network Configuration of Segmentation Network $\epsilon_s$}
\renewcommand\arraystretch{1.5}
\begin{tabular}{c|ccc}
\hline
Block & Layer & Channel / Kernel / Stride & Output size \\ \hline
Conv0 & Convolution* & 16\ /\ 3\ /\ 1 & 16$\times$224$\times$224 \\ \hline
\multirow{2}{*}{Contract1} & Max Pooling & -\ /\ 2\ /\ 2 & \multirow{2}{*}{32$\times$112$\times$112} \\
 & Convolution* & 32\ /\ 3\ /\ 1 & \\ \hline
\multirow{2}{*}{Contract2} & Max Pooling & -\ /\ 2\ /\ 2 & \multirow{2}{*}{64$\times$56$\times$56} \\
 & \multicolumn{1}{l}{Convolution*} & 64\ /\ 3\ /\ 1 & \\ \hline
\multirow{2}{*}{Contract3} & Max Pooling & -\ /\ 2\ /\ 2 & \multirow{2}{*}{\begin{tabular}[c]{@{}c@{}}128$\times$28$\times$28\\ ($s_1$)\end{tabular}} \\
 & \multicolumn{1}{l}{Convolution*} & 128\ /\ 3\ /\ 1 & \\ \hline
\multirow{2}{*}{Expand1} & Up Sampling & -\ /\ 2\ /\ - & \multirow{2}{*}{64$\times$56$\times$56} \\
 & \multicolumn{1}{l}{Convolution*} & 64\ /\ 3\ /\ 1 \\ \hline
\multirow{2}{*}{Expand2} & Up Sampling & -\ /\ 2\ /\ - & \multirow{2}{*}{32$\times$112$\times$112} \\
 & \multicolumn{1}{l}{Convolution*} & 32\ /\ 3\ /\ 1 & \\ \hline
\multirow{2}{*}{Expand3} & Up Sampling & -\ /\ 2\ /\ - & \multirow{2}{*}{\begin{tabular}[c]{@{}c@{}}16$\times$224$\times$224\\ ($f_s$)\end{tabular}} \\
 & \multicolumn{1}{l}{Convolution*} & 16\ /\ 3\ /\ 1 & \\ \hline
\multirow{2}{*}{SegOut} & \multirow{2}{*}{Convolution} & \multirow{2}{*}{1\ /\ 1\ /\ 1} & \multirow{2}{*}{\begin{tabular}[c]{@{}c@{}}1$\times$224$\times$224\\ ($M$)\end{tabular}} \\
& & & \\ \hline
\end{tabular}
\label{chap3:tbl:seg_unet}
\end{table}

The contracting path of $\epsilon_s$ also serves as a feature extractor for the authentication task. In our particular research context, finger vein authentication includes two distinct subtasks: verification and identification. Verification task compares feature embeddings of training and testing samples, while identification task determines the most likely class for the finger vein image based on class probabilities. The configuration of the subsequent authentication network $\epsilon_a$ is listed in Table \ref{chap2:tbl:resblock}. After segmentation block \textit{Contract3} outputs the feature map $s_1$, we employ two ResBlocks composed of Bottleneck layers in \cite{resnet} to further refine feature maps before applying the average pooling and linear layer to obtain the final category prior $\hat{y}_v$.
The goal of authentication task is to enlarge the distance of matching vectors between different classes, therefore Circle loss \cite{sun2020circle} is used to supervision this task:

{\small
\begin{equation}
\mathcal{L}_{auth} = log[1+
\sum^{L}_{j=1}exp(\gamma\alpha^{j}_{n}(s^{j}_{n}-\Delta_{n}))
\sum^{K}_{i=1}exp(-\gamma\alpha^{i}_{p}(s^{i}_{p}-\Delta_{p}))] 
\label{chap3:eq:circle}
\end{equation}
}
where $s_p$ and $s_n$ denote the intra-class and inter-class similarity scores, respectively; $\Delta_p$ and $\Delta_n$ are the intra-class and inter-class margins, respectively; $\alpha_p$ and $\alpha_n$ are non-negative weighting factors; $\gamma$ is a scale factor. In this paper, $\Delta_n$, $\Delta_p$ and $\gamma$ are set as 0.05, 0.95 and 128, respectively.

\begin{table}[!htbp]
\centering
\caption{Network Configuration of Authentication Network $\epsilon_a$}
\setlength{\tabcolsep}{5pt}
\renewcommand\arraystretch{1.5}
\begin{tabular}{c|ccc}
\hline
Block & Layer & Channel / Kernel / Stride & Output size \\ \hline
\multirow{3}{*}{ResBlock1} & \multirow{3}{*}{\begin{tabular}[c]{@{}c@{}}Bottleneck\\ $\times6$\end{tabular}} & 128\ /\ 1\ /\ 1 & \multirow{3}{*}{512$\times$14$\times$14} \\
& & 128\ /\ 3\ /\ 2 & \\
& & 512\ /\ 1\ /\ 1 & \\ \hline
\multirow{3}{*}{ResBlock2} & \multirow{3}{*}{\begin{tabular}[c]{@{}c@{}}Bottleneck\\ $\times3$\end{tabular}} & 256\ /\ 1\ /\ 1 & \multirow{3}{*}{1024$\times$7$\times$7} \\
& & 256\ /\ 3\ /\ 2 & \\
& & 1024\ /\ 1\ /\ 1 & \\ \hline
FeatOut & AveragePooling & -\ /\ 7\ /\ 1 & 1024$\times$1$\times$1 \\ \hline
DigitsOut & Linear & - & $N\ (\hat{y}_v)$ \\ \hline
\end{tabular}
\label{chap2:tbl:resblock}
\end{table}


\subsection{Diffusion Model and Denoising Branch} \label{chap3:sec:diffusion}
In the training stage, $\epsilon_a$ initially generates a category prior $\hat{y}_{v}$. Then, in the forward diffusion process, we add Gaussian noise to this vector based on the groundtruth one-hot label to obtain the noisy variable $y_t$. This process is mathematically expressed as follows:
\begin{align}
& y_t =\sqrt{\bar{\alpha}_t}y_0 + \sqrt{1-\bar{\alpha}_t}\epsilon + (1-\sqrt{\bar{\alpha}_t})\hat{y}_{v} \label{chap3:eq:diffusion} \\
with\quad & \bar{\alpha }_t=  {\textstyle \prod_{t}}\alpha_t, \alpha_t=1-\beta_t \nonumber
\end{align}
where $\epsilon$ denotes the Gaussian noise to be added, $\epsilon \sim \mathcal{N}(0,I)$; $\beta_t$ denotes the linear noise scheduling strategy and the noise is linearly scheduled with $\beta_1=1\times10^{-4}$ and $\beta_T=0.02$. The forward diffusion process is controlled by the time step $t$ sampled from a uniform distribution of $\left [ 1, T \right ] $, yielding a sequence of noisy variables denoted as $\left \{ y_1, y_2, \cdots, y_T \right \}$.

Subsequently, we concatenate the noisy variable $y_t$ with the prior $\hat{y}_{v}$ before feeding this combination into the U-Net-like denoising model $\epsilon_d$. To improve the model's proficiency in learning the noise distribution, we developed a conditional guidance called mask condition to the denoising network. This module allowed the denoising network to perceive vein pattern semantics from the feature maps in the segmentation network. Given that the segmentation network's outputs are 2-dimensional features, which are incompatible with the denoising network's vectorized features, the mask condition should also take care of dimensionality reduction. The detailed operations of mask condition is elaborated as:
\begin{align}
& f_{c} = Max(f_{s}*k_{Gauss}, f_s) \\
& f_{c}^{\prime} = M_c(f_{c}) \otimes f_{c}, \ f_{c}^{\prime \prime} = M_c(f_{c}^{\prime}) \otimes f_{c}^{\prime} \\
& f_{d}^{\prime} = MaxPool(f_{c}^{\prime \prime}) * k_{Conv1\times1} + f_{d}
\label{chap2:eq:condition}
\end{align}
where: $\otimes$ denotes element-wise multiplication;
$M_{c}(\cdot)$ and $M_{s}(\cdot)$ denotes the channel-wise attention and spatial-wise attention as proposed in \cite{woo2018cbam}, respectively.
Specifically, we first apply a learnable Gaussian kernel $k_{Gauss}$ to smooth the segmented feature map. The smoothed feature map is then element-wise maximized with the original one to obtain the enhanced features $f_{c}$ by preserving the most prominent information
Then CBAM\cite{woo2018cbam} is used to obtain enhanced features and the following max pooling further reduces the scale of the feature map.
The feature map is further condensed via 1$\times$1 convolution to reduce channels before integration into the initial denoising network layer, yielding the semantically enriched embedding $f_{d}^{\prime}$. The noise prediction by the denoising network can be represented mathematically as:
\begin{equation}
\hat{\epsilon} = \epsilon_d(f_{d}^{\prime}, y_t, \hat{y}_v, t) = D(E(f([y_t, \hat{y}_v]), f_{d}^{\prime}, t), t)
\label{chap2:eq:denoisenet}
\end{equation}
where: $f(\cdot)$ denotes the linear projection to the latent space; $[\cdot]$ denotes the concatenate operation; $E(\cdot)$ and $D(\cdot)$ denote the encoder and decoder of the denoising network, respectively; $f_{d}^{\prime}$ denotes the mask condition from the segmentation network. 
In each layer of the denoising network, we use Hadamard product to fuse the feature embedding with the timestep embedding. The fused embeddings are fed into two consecutive fully connected (FC) layers for noise prediction. Each layer first generates a feature embedding, which is then element-wise multiplied with the timestep embedding. 


In the inference stage, the denoising network operates as a parameterized inverse diffusion process. The trained $\epsilon_d$ effectively converts the noisy variable distribution $p_{\theta}(y_T)$ back to the original clean data distribution $p_{\theta}(y_0)$. This transformation is governed by the equation detailed in Eq. \ref{chap2:eq:reparam}. As a result of this process, the network produces the final prediction $\hat{y}_0$, which approximates the groundtruth.
\begin{align}
& p_{\theta}(y_T)=\mathcal{N}(\hat{y}_v, I) \\
& p_{\theta}(y_{0:T-1}|y_{T},f_{d}^{\prime})=\prod _{t=1}^{T}p_{\theta}(y_{t-1}|y_{t}, f_{d}^{\prime}) \label{chap2:eq:reparam}
\end{align}
where $\theta$ denotes the parameters of the denoising UNet; $\mathcal{N}(\cdot, \cdot)$ denotes the Gaussian distribution; and $I$ is the identity matrix.

To augment the denoising network's ability to accurately predict noise with specific spectral characteristics, we introduce the Fourier-space Structural Similarity (FourierSIM) loss. This novel loss function is inspired by the widely recognized structural similarity index (SSIM), which has been a benchmark in image restoration tasks. SSIM evaluates the perceptual quality of images by comparing their luminance, contrast, and structures, which are quantified through the mean, variance, and covariance of the image pairs. The SSIM loss derived from this index is formulated as:
\begin{align}
& SSIM(x,y) = \frac{(2\mu_x\mu_y+C_1)(2\sigma_{xy}+C_2)}{(\mu_x^2 + \mu_y^2 + C_1)(\sigma_x^2 + \sigma_y^2+C2)} \\
& L_{SSIM}(x, y) = 1 - SSIM(x, y)
\end{align}
where $x$ and $y$ represents the image pair; $\mu_{x,y}$, $\sigma_{x,y}$ and $\sigma_{xy}$ represents the mean value, standard deviation and covariance, respectively. 
By extending this concept into the Fourier domain, our FourierSIM loss assesses similar attributes within the spectral components of noise. This allows for a more nuanced understanding of the noise patterns and their structural integrity. Specifically, we take into account the discrepancy of amplitude and phase between the groundtruth and predicted noise and yield:
\begin{align}
L_{FourierSIM} = L_{SSIM}(A(\epsilon), A(\hat{\epsilon})) + L_{SSIM}(\phi(\epsilon), \phi(\hat{\epsilon}))
\end{align}
where $\epsilon$ and $\hat{\epsilon}$ represents the groundtruth and predicted noise, respectively; $A(\cdot)$ and $\phi(\cdot)$ denotes the amplitude and phase of noise in Fourier space, respectively. Additionally, we incorporate the mean square error (MSE) loss as a complementary measure as is commonly practiced in other DDPMs. To summarize, the complete loss of the denoising task in this paper can be formulated as:
\begin{equation}
\mathcal{L}_{diff} = \mathcal{L}_{MSE} + \lambda \mathcal{L}_{FourierSIM} 
\end{equation}
where $\lambda$ is the weight factor and it is set as 0.5 in this paper.

\subsection{Semantic Difference Transformer} \label{chap3:sec:sdformer}

The authentication task in this paper employs a classification diffusion model, which inherently relies on category label vectors for its diffusion and denoising processes. Therefore, we hold that the denoising network harbours latent category embeddings within its intermediate feature embeddings. These embeddings are particularly crucial for the segmentation task, as they provide the discriminative information required for precise vein pattern extraction and bridge the gap of our current segmentation network.
Moreover, the intermediate features of the denoising network are shaped by both the diffusion model and conditional priors from  segmentation encoder. This means that enriching segmentation feature representations could potentially benefit the authentication task.

To this end, we propose the Semantic Difference Transformer (SD-Former), a feature fusion module that introduces latent denoising category embeddings to the segmentation network. This approach has advantages over directly employing category labels, facilitating the integration of segmentation and authentication into a unified learning framework.
\begin{figure}[htbp]
\centering
\begin{minipage}{0.8\linewidth}\centering
\includegraphics[width=0.9\textwidth]{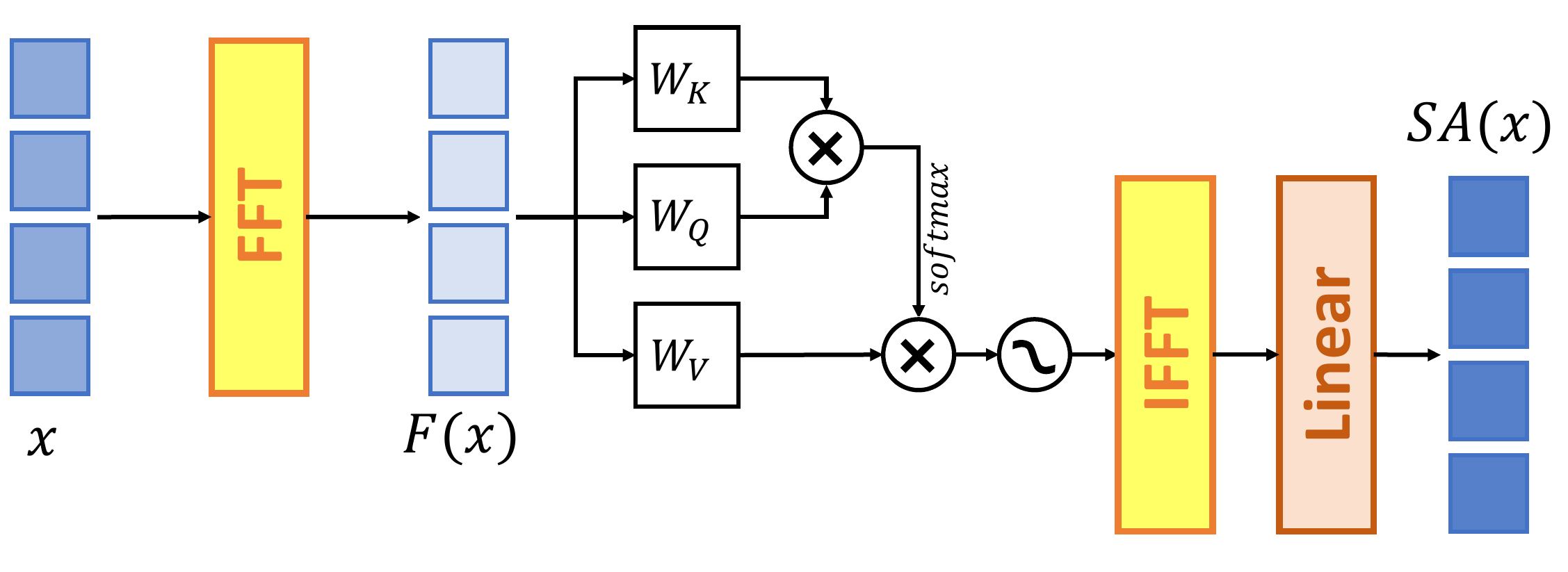}
\centerline{\footnotesize{(a) Fourier-space Self-Attention (FSA)}}
\end{minipage}
\begin{minipage}{0.8\linewidth}\centering
\includegraphics[width=0.95\textwidth]{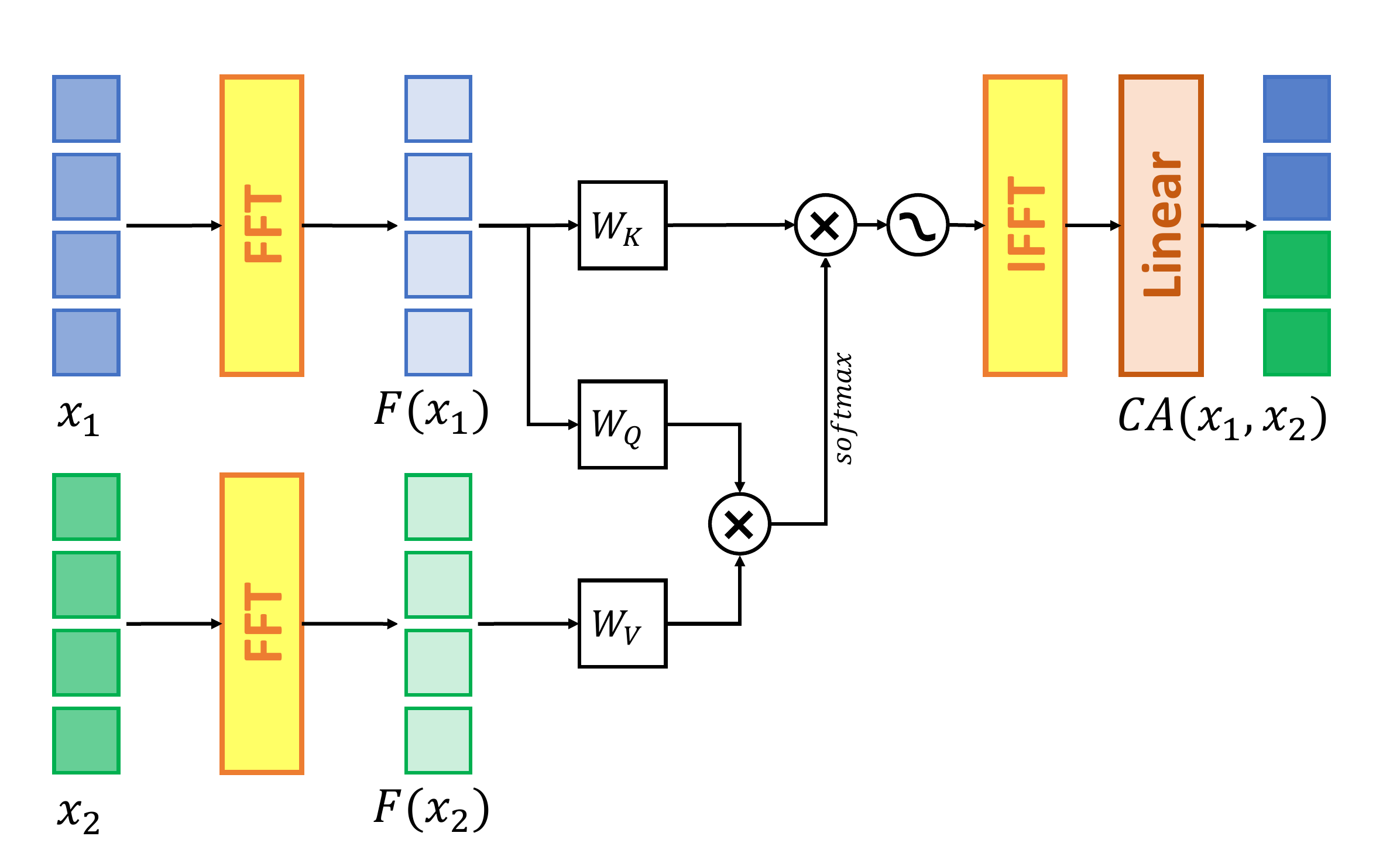}
\centerline{\footnotesize{(b) Fourier-space Cross-Attention (FCA)}}
\end{minipage}
\caption{Illustration of Fourier-space attention modules.}
\label{chap3:fig:attention}
\end{figure}

However, a notable challenge is the domain discrepancy between the diffusion embeddings and the segmentation feature maps and directly fusing diffusion embeddings into the segmentation network can be problematic. To address this, we draw on the concepts from MedSegDiff \cite{wu2023medsegdiff} and developed the Fourier-space Self-Attention (FSA) and Fourier-space Cross-Attention (FCA) modules, as depicted in Fig. \ref{chap3:fig:attention}.

The Fourier Transform is instrumental in revealing the statistical properties of feature embeddings across various frequencies, which is particularly beneficial for denoising-related tasks. The FSA module distinguishes semantic information (such as category and noise) by frequency characteristics within the embeddings, while the FCA module semantically aligns components with similar frequency characteristics between two sets of embeddings. The algorithms for FSA and FCA are elaborated in Eq. (\ref{chap2:eq:fsa})-(\ref{chap2:eq:fsa}), respectively. Both modules implement multi-head attention in Fourier space. The sine function with learnable parameters serve as the activation function to constrain the spectrum in Fourier space. These features are then aggregated through linear projection after inverse Fourier Transform.
\begin{align}
& Attn_{SA} = Softmax((\mathcal{F}(x)W_Q)(\mathcal{F}(x)W_K)^T)(\mathcal{F}(x)W_V) \nonumber \\ 
& M_{SA}= a\ sin( \omega Attn_{SA}) \nonumber \\
& f_{SA}=\mathcal{F}^{-1}(M_{SA}) W_{SA}\label{chap2:eq:fsa} \\
& Attn_{CA} = Softmax((\mathcal{F}(x_2)W_Q)(\mathcal{F}(x_1)W_K)^T)(\mathcal{F}(x_1)W_V) \nonumber \\ 
& M_{CA}=a\ sin( \omega Attn_{CA} ) \nonumber \\
& f_{CA}=\mathcal{F}^{-1}(M_{CA}) W_{CA} \label{chap2:eq:fca}
\end{align}
where $W_{Q,K,V}$ denote the projection matrices for \textit{query}, \textit{key} and \textit{value}; $\mathcal{F}(\cdot)$ and $\mathcal{F}^{-1}(\cdot)$ denote the Fourier Transform and its inverse; $a$ and $\omega$ are learnable amplitude and frequency parameters; $W_{SA}$ and $W_{CA}$ are linear layers used for aggregating features in FSA and FCA, respectively.

\begin{figure}[htbp]
\centering
\begin{minipage}{0.8\linewidth}\centering
\includegraphics[width=0.9\textwidth]{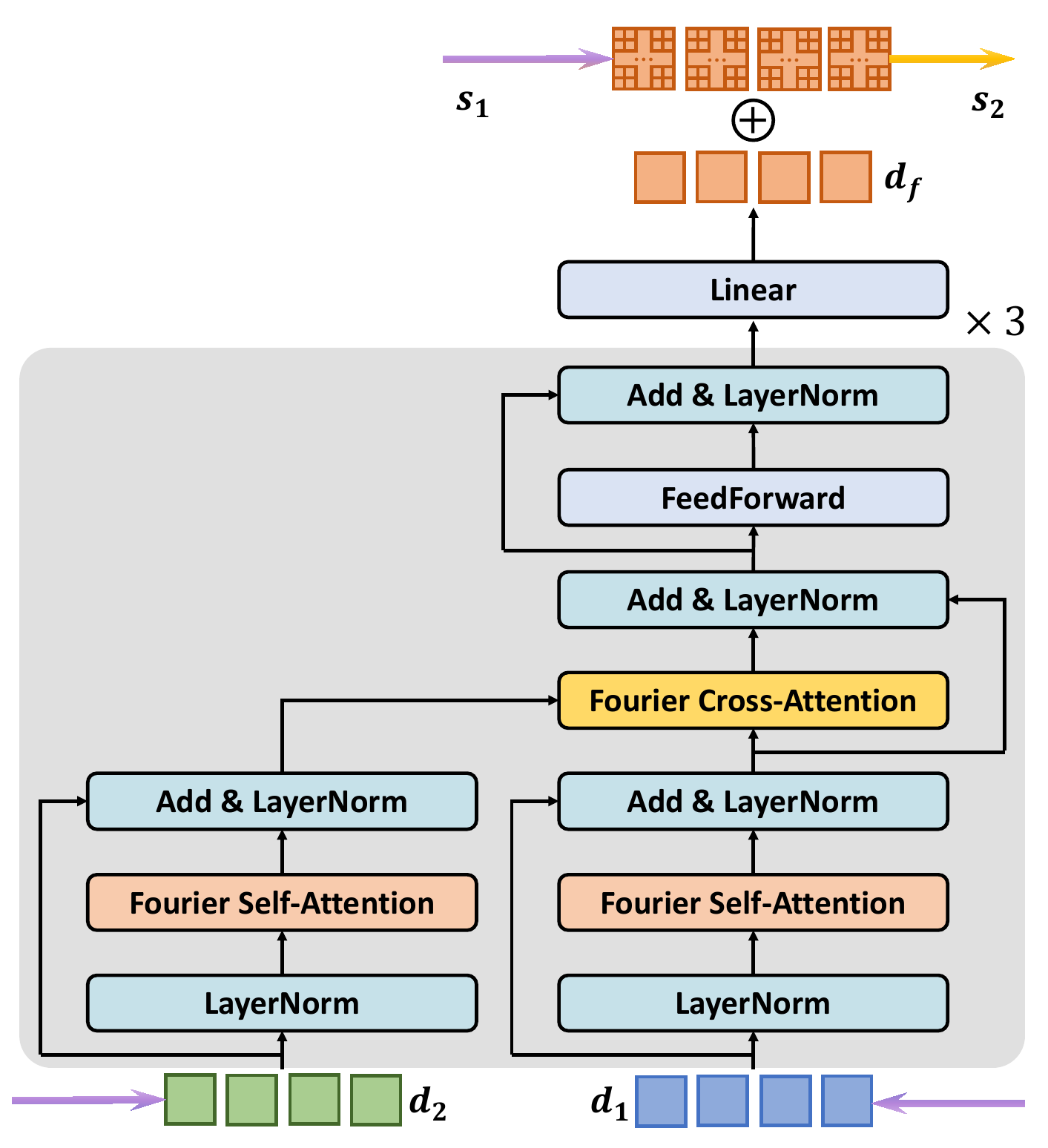}
\end{minipage}
\caption{Illustration of SD-Former architecture.}
\label{chap3:fig:former}
\end{figure}

The SD-Former's architecture, shown in Fig. \ref{chap3:fig:former}, draws inspiration from the classic Transformer design and consists of several identical and consecutive blocks (three in our setting). Each of the block consists of FSA, FCA, feedforward layer, shortcut connections and layer normalization. In our implementation, SD-Former takes as input the feature embedding $d_1$ and $d_2$ from the initial and final layers of the denoising network, respectively. The resulting fused embedding $d_f$ is added to the feature map $s_1$ from the contracting path of the segmentation network to produce a new feature map $s_2$, which is then input to the expanding path of the segmentation network.

\section{Experiments}
In this section, extensive experiments are conducted on the USM and THU-MVFV3V datasets. Both segmentation and authentication results are provided for the comprehensive analysis of our proposed method. We elaborate the details of datasets and the generation of groundtruth vein masks in the Sec. \ref{chap4:sec:dataset}. Our implementation details and evaluation metrics are provided in Sec. \ref{chap4:sec:settings}. Comparative experiments and analysis are presented in Sec. \ref{chap4:sec:auth} and \ref{chap4:sec:seg}. Also, we designed a series of ablation experiments to further study the effectiveness of our proposed modules, which can be seen in Section \ref{chap4:sec:ablation}.
\begin{table}[htbp]
\centering
\caption{Details of Finger Vein Datasets}
\renewcommand\arraystretch{1.5}
\begin{tabular}{l|cc}
\hline
Dataset & USM \cite{asaari2014fusion} & THU-MVFV3V \cite{zhao2023vpcformer} \\ \hline
Finger type & Middle \& Index & Middle \& Index \\ \hline
ROI size & 300$\times$100 & 300$\times$150 \\ \hline
Subject Category & 492 & 660 \\ \hline
Samples per Finger & 6 & 6 \\ \hline
Total Images in use & 5904 & 7920 \\ \hline
\end{tabular}
\label{chap4:tbl:datasets}
\end{table}
\subsection{Datasets \label{chap4:sec:dataset}}
\subsubsection{USM Dataset}
\begin{figure}[htbp]
\centering
\begin{minipage}{0.4\linewidth}\centering
\includegraphics[width=\linewidth]{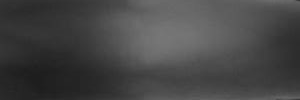}
\centerline{(a)}
\end{minipage}
\hspace{5pt}
\begin{minipage}{0.4\linewidth}\centering
\includegraphics[width=\linewidth]{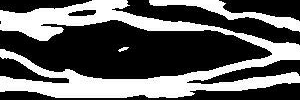}
\centerline{(b)}
\end{minipage}
\caption{A sample of USM dataset. (a) ROI image; (b) Vein mask.}
\label{chap4:fig:usm_dataset}
\end{figure}

USM \cite{asaari2014fusion} dataset was constructed by the research team from Universiti Sains Malaysia in 2014. This dataset includes a collection of finger vein images from 123 volunteers and each of them provided their index and middle fingers of both hands. Each finger was captured for 6 times in two separate sessions with a time interval of more than two weeks. Therefore, there are 492 finger categories with a total of 5904 finger vein images in this dataset. The resolution of raw images is 640$\times$480 pixels and a region of interest (ROI) with the size of $300\times100$ pixels is also provided for each raw finger vein image, as illustrated in Fig. \ref{chap4:fig:usm_dataset} (a).

In order to supervise the segmentation task, the mask generation module (MGM) in EIFNet \cite{song2022eifnet} was used here to generate groundtruth masks. As advised in EIFNet, we trained the MGM on the THU-FVS dataset with human-annotated vein masks and followed the default settings to generate the vein masks of USM dataset. An example vein mask is presented in Fig. \ref{chap4:fig:usm_dataset} (b).

\subsubsection{THU-MVFV3V Dataset}
THU-MVFV3V \cite{zhao2023vpcformer} dataset was established by Tsinghua University in 2023. The images in this dataset were collected from 92 males and 79 females. Each volunteer provided their index and middle fingers of both hands and the total number of finger categories is 660 after leaving out poorly-captured samples. Similarly, the data acquisition process included two phases with a time span of 45.8 days in average. Each finger was sampled 6 times at three angles (-45$^{\circ}$, 0$^{\circ}$, 45$^{\circ}$) in each session, resulting in a total of 23,760 finger vein images. It should be noted that only the images taken at the frontal angle (0$^{\circ}$) are used in our work, which means totally 7920 images are used in the following experiments. A $300\times150$ pixels ROI is also provided for each image.

\begin{figure}[htbp]
\centering
\begin{minipage}{0.15\linewidth}\centering
\includegraphics[width=\linewidth]{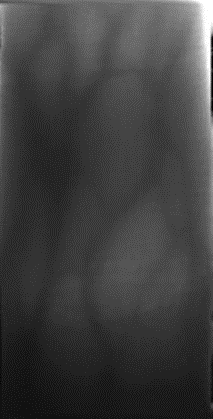}
\centerline{(a)}
\end{minipage}
\begin{minipage}{0.15\linewidth}\centering
\includegraphics[width=\linewidth]{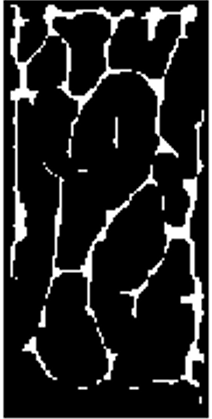}
\centerline{(b)}
\end{minipage}
\begin{minipage}{0.15\linewidth}\centering
\includegraphics[width=\linewidth]{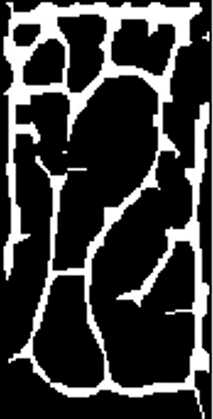}
\centerline{(c)}
\end{minipage}
\begin{minipage}{0.15\linewidth}\centering
\includegraphics[width=\linewidth]{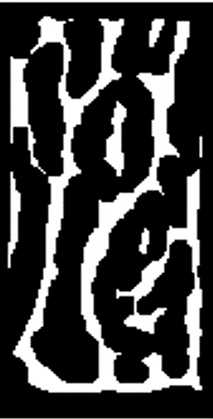}
\centerline{(d)}
\end{minipage}
\begin{minipage}{0.15\linewidth}\centering
\includegraphics[width=\linewidth]{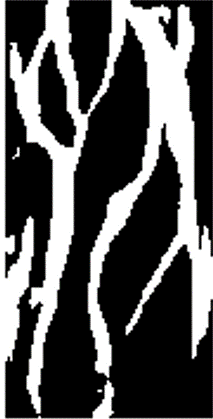}
\centerline{(e)}
\end{minipage}
\begin{minipage}{0.15\linewidth}\centering
\includegraphics[width=\linewidth]{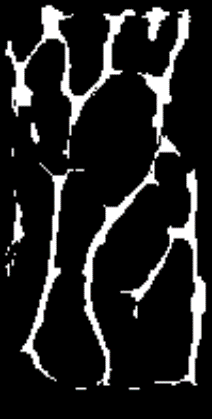}
\centerline{(f)}
\end{minipage}
\caption{A sample of THU-MVFV3V dataset. (a) ROI image; (b) Vein mask by LMC; (c) Vein mask by EMC; (d) Vein mask by Gabor filter; (e) Vein mask by EIFNet; (f) Final vein mask.}
\label{chap4:fig:THUMVFV3V_dataset}
\end{figure}

Since no human correlation was involved in the learning-based vein mask generation, we combined the segmentation results of EIFNet with those of another three explicit vein extraction methods LMC \cite{miura2007extraction}, EMC \cite{syarif2017enhanced}, Gabor filter \cite{kumar2011human}. We adopted the majority vote algorithm (MVA) to generate the vein masks for THU-MVFV3V dataset. Firstly, each of the algorithms generate their predicted masks separately. When a pixel is labeled as a vein point in more than three predicted masks, then this pixel can be regarded as a reliable vein point. Otherwise, this pixel should be labelled as background. Fig. \ref{chap4:fig:THUMVFV3V_dataset} shows the vein masks predicted by different methods and the final result by MVA.
\subsection{Experimental Settings \label{chap4:sec:settings}}
\subsubsection{Implementation Details}
Our framework is implemented with PyTorch on a single NVIDIA RTX 3090 GPU. Before the training process, we randomly cropped a square area on each ROI image and its corresponding binary mask. Then we resize their resolution to $224\times224$ pixels to adapt to our network. Random change in contrast and Gaussian blur are applied on the training images for data augmentation.

In the training phase, the initial learning rate for the segmentation and denoising networks are set as $1\times10^{-4}$ and $1\times10^{-3}$ respectively. The entire framework is trained in an end-to-end manner using the AdamW optimizer. It should be noted that following the training scheme in \cite{yang2023diffmic}, the segmentation branch needs to be pre-trained at first to provide a preliminary prediction before training the entire framework jointly. The pre-training epoch is set as 50 epochs and the number of total training epochs is set as 500 for both datasets. The batch size is set as 4 for both pre-training and joint-training. Following the common practice in this field \cite{song2022eifnet,zhao2023vpcformer}, we split the dataset into the training and testing set according to the data acquisition session. Images in session one is used for training while images in session two is used for evaluation. 

In the inference phase, based on the empirical setting in \cite{yang2023diffmic} and our experimental results, we set the total diffusion time steps $T$ as 100 for USM dataset and 200 for THU-MVFV3V dataset.

\subsubsection{Evaluation Metrics}
To evaluate the segmentation and authentication performance of our proposed method, we adopted several mainstream metrics for fair comparison with other popular methods. Both verification and identification tasks are included in finger vein authentication.

EER, as a widely used metrics in biometric recognition, is used here to evaluate the matching performance between the features of training and testing samples, which we term as \textit{finger vein verification} task in this paper. In the detection error tradeoff (DET) \cite{martin1997det} curve, EER represents the intersection where false acceptance rate (FAR) coincides with false rejection rate (FRR). We provide both DET curve and EER value in the following experiments for comprehensive analysis. 

ACC is a commonly-used metrics for classification tasks and it is used here to measure the matching performance between predicted and groundtruth category, which we term as \textit{finger vein identification} task. Different from EER, ACC only evaluate the accuracy of predicted finger categories in the testing test. 

For finger vein segmentation, we used a series of volumetric scores including mean Dice coefficient (Dice), centerline Dice (clDice) \cite{shit2021cldice}, accuracy (ACC) and AUC to evaluate the overall segmentation performance. It is important to note that clDice is particularly representative here because it assesses topological continuity, which aligns well with the inherent tubular structure of finger veins.

\subsection{Finger Vein Authentication \label{chap4:sec:auth}}
In this section, we compared the authentication performance of DiffVein with those of several widely-used traditional and deep learning-based methods on USM and THU-MVFV3V datasets.

\begin{figure}[!htbp]
\centering
\includegraphics[width=0.85\linewidth]{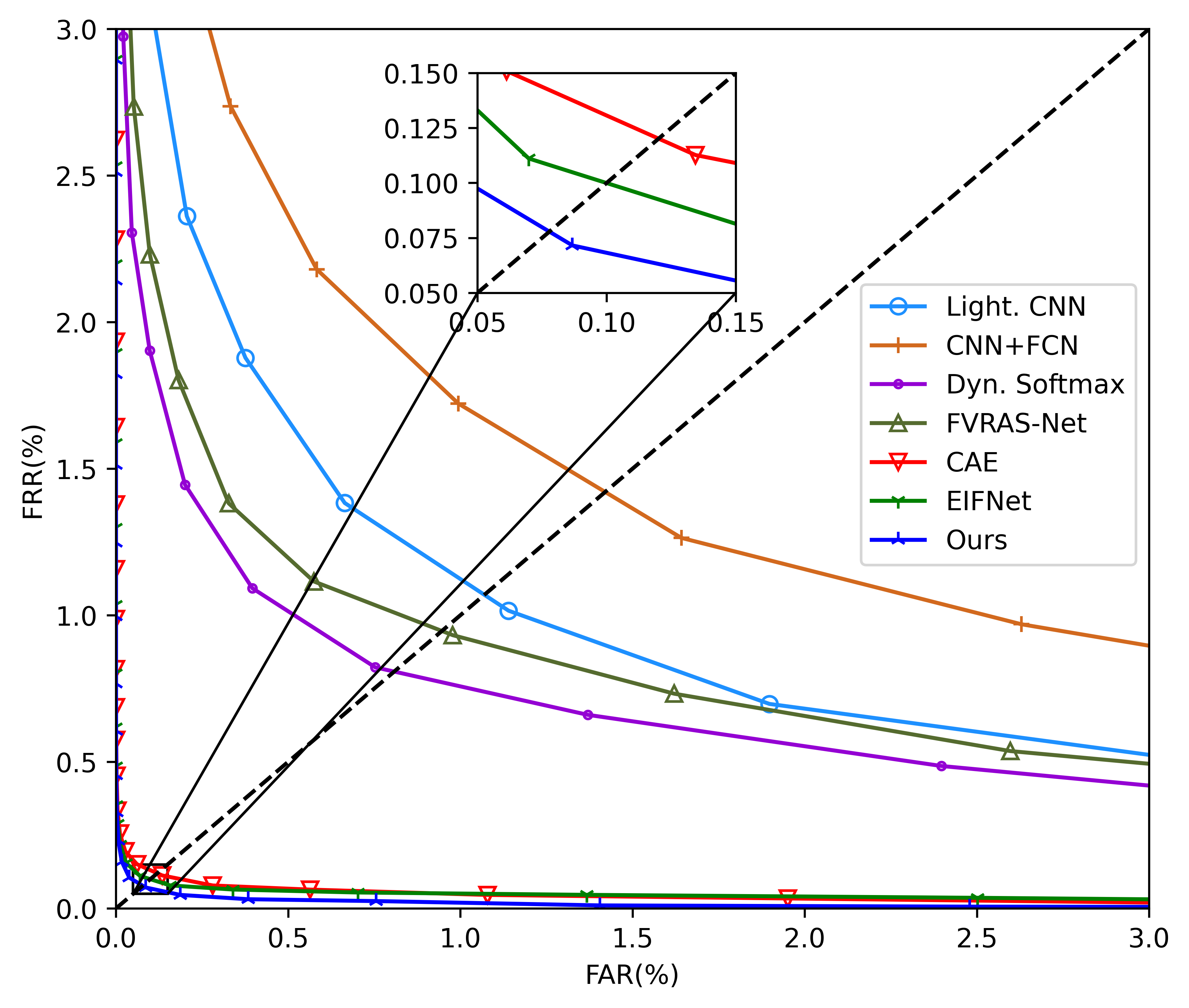}
\caption{DET curve of different methods on USM dataset.}
\label{chap4:fig:usm_auth}
\end{figure}

\begin{figure}[!htbp]
\centering
\includegraphics[width=0.85\linewidth]{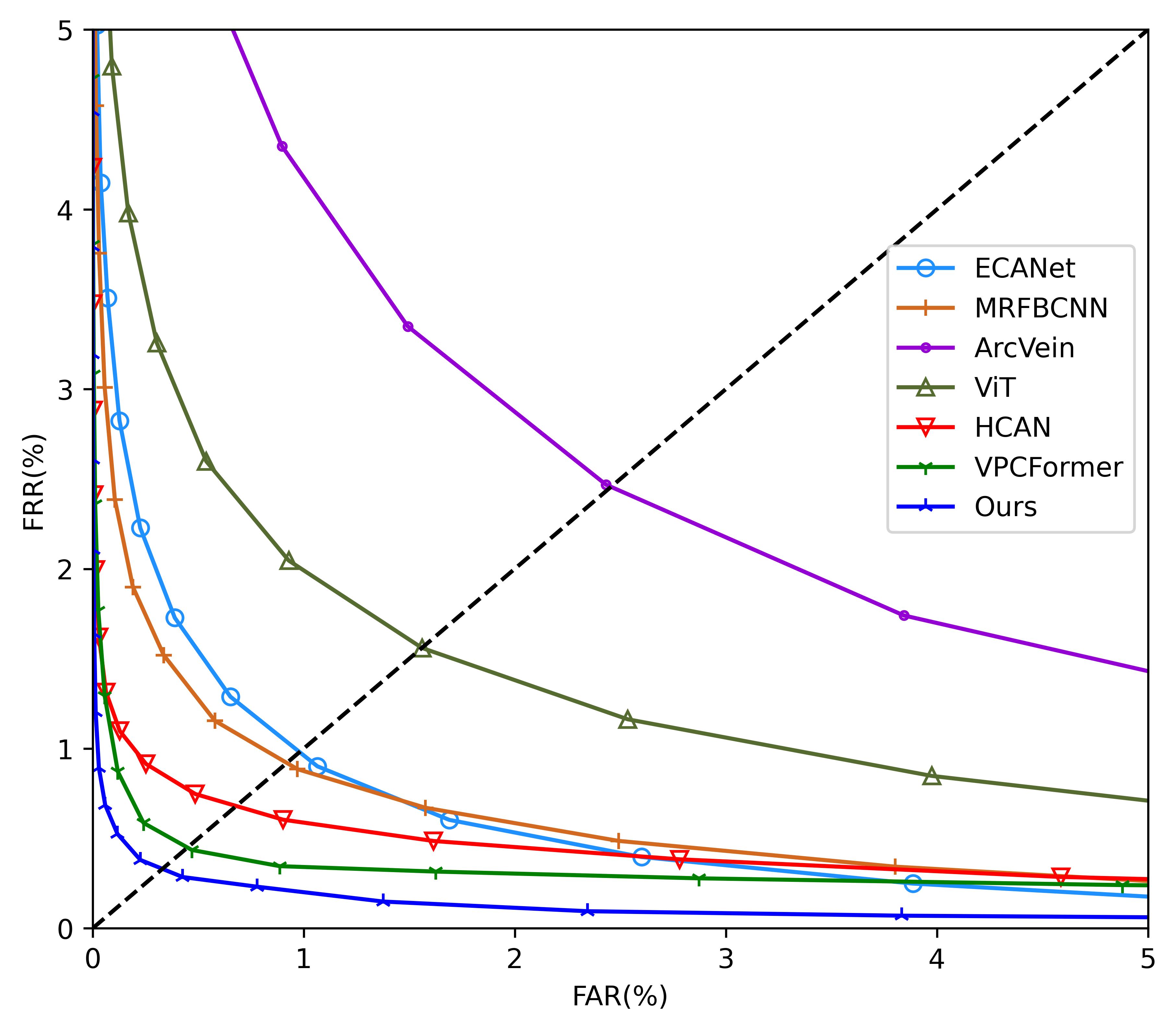}
\caption{DET curve of different methods on THU-MVFV3V dataset.}
\label{chap4:fig:thumvfv3v_auth}
\end{figure}

The DET curves of comparative experiments on both datasets are shown in Fig. \ref{chap4:fig:usm_auth} and Fig. \ref{chap4:fig:thumvfv3v_auth}. It can be seen from Fig. \ref{chap4:fig:usm_auth} that CAE \cite{hou2019convolutional}, EIFNet \cite{song2022eifnet} and our DiffVein significantly outperform other methods in verification task and our method slightly leads the other two opponents. THU-MVFV3V dataset is more complicated than USM so the distinction of performance between different methods is more remarkable. As can be observed in Fig. \ref{chap4:fig:thumvfv3v_auth}, the DET curve of our method is consistently lower than that of VPCFormer \cite{zhao2023vpcformer} and HCAN \cite{zhao2022exploiting} proposed in recent years, which indicates our method outperform the other existing methods.

More quantitative results are shown in Table \ref{chap4:tbl:usm_auth} and Table \ref{chap4:tbl:thumvfv3v_auth}. The best results are highlighted in bold. It can be observed that learning-based methods generally performs better than traditional methos under the EER and ACC metrics on both of the datasets. It means that these methods are prone to predict more consistent intra-class features and more accurate inter-class features, which validates the powerful capacity of deep networks in extracting discriminative patterns. Our DiffVein shows the leading performance across all the other methods in both verification and identification tasks. Our explanation for the improved performance is that conditioning the denoising network with segmentation feature maps facilitates easier convergence, thereby enhancing its effectiveness. The further ablation analysis on the effectiveness of diffusion model and mask condition is provided in Sec. \ref{chap4:sec:ablation}.
\begin{table}[htbp]
\centering
\caption{Comparison of Authentication Results of Different Methods on USM dataset}
\renewcommand\arraystretch{1.5}
\setlength{\tabcolsep}{0.03\columnwidth}
\begin{tabular}{c|l|cc}
\hline
Type & Method & EER(\%) & ACC(\%)\\ \hline
\multirow{8}{*}{\begin{tabular}[c]{@{}c@{}}Traditional\\ methods\end{tabular}}  
& RLT \cite{miura2004feature} & 6.73 &  92.04 \\ 
& LMC \cite{miura2007extraction} & 4.04 &  97.01 \\
& PC \cite{choi2009finger} & 4.34 &  96.87 \\
& WLD \cite{huang2010finger} & 4.41 &  96.55 \\
& Gabor filters \cite{kumar2011human} & 3.47 &  97.32 \\
& EMC \cite{syarif2017enhanced} & 1.63 &  98.42 \\
& ASAVE \cite{yang2017finger} & 5.14 &  95.01 \\
& LBP \cite{lee2009finger} & 5.76 &  95.48 \\ \hline
\multirow{7}{*}{\begin{tabular}[c]{@{}c@{}}Learning-based\\ methods\end{tabular}} 
& CNN+FCN \cite{qin2017deep} & 1.42 &  98.41 \\
& CAE \cite{hou2019convolutional} & 0.12 &  99.54 \\
& FVRAS-Net \cite{yang2020fvras} & 0.95 &  98.81 \\
& Lightweight CNN \cite{zhao2020finger} & 1.07 &  98.53 \\
& Dynamic softmax \cite{li2022improving} & 0.81 &  98.79 \\
& EIFNet \cite{song2022eifnet} & 0.10 &  99.76 \\
& \textbf{Ours}   & \textbf{0.08} & {\textbf{99.79}}  \\ \hline
\end{tabular}
\label{chap4:tbl:usm_auth}
\end{table}

\begin{table}[htbp]
\centering
\caption{Comparison of Authentication Results of Different Methods on THU-MVFV3V dataset}
\renewcommand\arraystretch{1.5}
\setlength{\tabcolsep}{0.04\columnwidth}
\begin{tabular}{c|l|cc}
\hline
Type & Method & EER(\%) & ACC(\%) \\ \hline
\multirow{5}{*}{\begin{tabular}[c]{@{}c@{}}Traditional\\ methods\end{tabular}} 
& LMC \cite{miura2007extraction} & 7.65  &  91.13 \\
& WLD \cite{huang2010finger} & 5.64  &  95.51 \\
& LDC \cite{meng2012finger} & 6.05  &  92.73 \\
& Gabor filters \cite{kumar2011human} & 8.68  &  90.88 \\
& PWBDC \cite{yang2019hybrid} & 7.56 &  91.03 \\ \hline
\multirow{8}{*}{\begin{tabular}[c]{@{}c@{}}Learning-based\\ methods\end{tabular}} 
& ResNet50 \cite{resnet} & 1.23  & 98.23 \\
& ECANet \cite{wang2020eca} & 0.98  & 98.82 \\
& MRFBCNN \cite{wang2021finger} & 0.92  & 98.76 \\
& ArcVein \cite{hou2021arcvein} & 2.45  & 96.97 \\
& ViT \cite{dosovitskiy2020vit} & 1.56  & 98.89 \\
& HCAN \cite{zhao2022exploiting} & 0.68  & 99.44 \\
& VPCFormer \cite{zhao2023vpcformer} & 0.45 & 99.67 \\
& \textbf{Ours} & \textbf{0.33} & \textbf{99.71}    \\ \hline
\end{tabular}
\label{chap4:tbl:thumvfv3v_auth}
\end{table}
\subsection{Finger Vein Segmentation \label{chap4:sec:seg}}
In this section, we compared our DiffVein with several general segmentation methods including U-Net \cite{cciccek20163d}, TransUNet \cite{chen2021transunet} and methods designed especially for tubular structure segmentation including DCU-Net \cite{yang2022dcu}. It should be noted that while the ground truth vein masks used in our paper do not accurately represent the location and geometry of all finger veins, they still serve as a valuable guide for segmentation tasks aimed at vein feature extraction. More precisely, our goal is to generate more meaningful and representative vein masks that can improve the finger vein authentication process.

The volumetric scores are listed in Table \ref{chap4:tbl:all_seg} and the best results are highlighted in bold. It can be seen from the table that our segmentation network clearly outperform other methods in all of the volumetric scores on both datasets. 
The segmentation network in DiffVein demonstrated its effectiveness in preserving the veins' topology, as evidenced by the clDice coefficients of 93.30\% for the USM dataset and 93.16\% for the THU-MVFV3V dataset.

\begin{figure*}[htbp]
\centering
\begin{minipage}{0.16\linewidth}\centering
\includegraphics[width=0.95\linewidth]{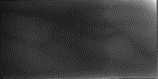}
\centerline{\quad}
\end{minipage}
\begin{minipage}{0.16\linewidth}\centering
\includegraphics[width=0.95\linewidth]{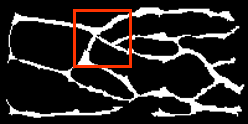}
\centerline{\quad}
\end{minipage}
\begin{minipage}{0.16\linewidth}\centering
\includegraphics[width=0.95\linewidth]{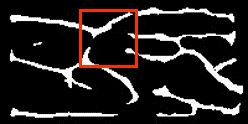}
\centerline{\quad}
\end{minipage}
\begin{minipage}{0.16\linewidth}\centering
\includegraphics[width=0.95\linewidth]{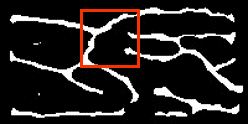}
\centerline{\quad}
\end{minipage}
\begin{minipage}{0.16\linewidth}\centering
\includegraphics[width=0.95\linewidth]{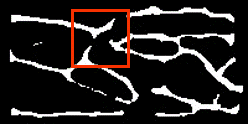}
\centerline{\quad}
\end{minipage}
\begin{minipage}{0.16\linewidth}\centering
\includegraphics[width=0.95\linewidth]{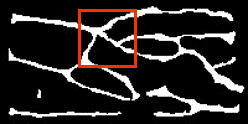}
\centerline{\quad}
\end{minipage}
\begin{minipage}{0.16\linewidth}\centering
\includegraphics[height=0.055\textheight]{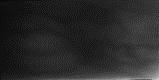}
\centerline{(a)}
\end{minipage}
\begin{minipage}{0.16\linewidth}\centering
\includegraphics[height=0.055\textheight]{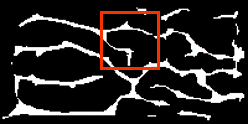}
\centerline{(b)}
\end{minipage}
\begin{minipage}{0.16\linewidth}\centering
\includegraphics[height=0.055\textheight]{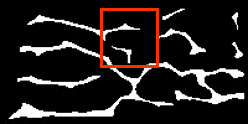}
\centerline{(c)}
\end{minipage}
\begin{minipage}{0.16\linewidth}\centering
\includegraphics[height=0.055\textheight]{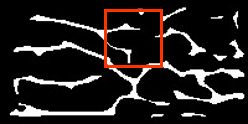}
\centerline{(d)}
\end{minipage}
\begin{minipage}{0.16\linewidth}\centering
\includegraphics[height=0.055\textheight]{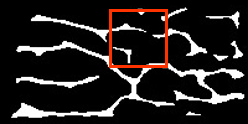}
\centerline{(e)}
\end{minipage}
\begin{minipage}{0.16\linewidth}\centering
\includegraphics[height=0.055\textheight]{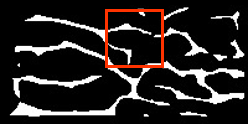}
\centerline{(f)}
\end{minipage}
\caption{Visualization of finger vein segmentation results on THU-MVFV3V dataset using different methods: (a) ROI image; (b) Groundtruth mask; (c) UNet; (d) TransUNet; (e) DCUNet; (f) Ours.}
\label{chap4:fig:thumvfv3v_seg}
\end{figure*}

\begin{figure*}[htbp]
\centering
\begin{minipage}{0.16\linewidth}\centering
\includegraphics[width=0.95\linewidth]{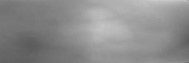}
\centerline{\quad}
\end{minipage}
\begin{minipage}{0.16\linewidth}\centering
\includegraphics[width=0.95\linewidth]{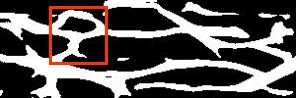}
\centerline{\quad}
\end{minipage}
\begin{minipage}{0.16\linewidth}\centering
\includegraphics[width=0.95\linewidth]{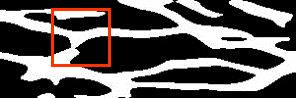}
\centerline{\quad}
\end{minipage}
\begin{minipage}{0.16\linewidth}\centering
\includegraphics[width=0.95\linewidth]{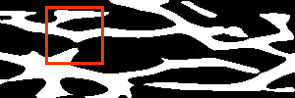}
\centerline{\quad}
\end{minipage}
\begin{minipage}{0.16\linewidth}\centering
\includegraphics[width=0.95\linewidth]{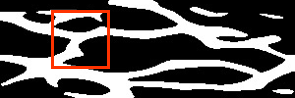}
\centerline{\quad}
\end{minipage}
\begin{minipage}{0.16\linewidth}\centering
\includegraphics[width=0.95\linewidth]{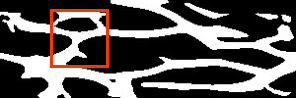}
\centerline{\quad}
\end{minipage}
\begin{minipage}{0.16\linewidth}\centering
\includegraphics[width=0.95\linewidth]{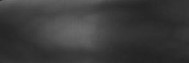}
\centerline{(a)}
\end{minipage}
\begin{minipage}{0.16\linewidth}\centering
\includegraphics[width=0.95\linewidth]{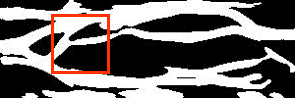}
\centerline{(b)}
\end{minipage}
\begin{minipage}{0.16\linewidth}\centering
\includegraphics[width=0.95\linewidth]{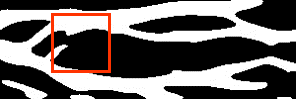}
\centerline{(c)}
\end{minipage}
\begin{minipage}{0.16\linewidth}\centering
\includegraphics[width=0.95\linewidth]{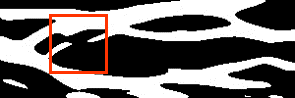}
\centerline{(d)}
\end{minipage}
\begin{minipage}{0.16\linewidth}\centering
\includegraphics[width=0.95\linewidth]{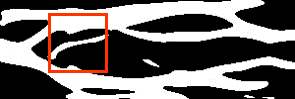}
\centerline{(e)}
\end{minipage}
\begin{minipage}{0.16\linewidth}\centering
\includegraphics[width=0.95\linewidth]{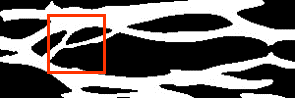}
\centerline{(f)}
\end{minipage}
\caption{Visualization of finger vein segmentation results on USM dataset using different methods: (a) ROI image; (b) Groundtruth mask; (c) UNet; (d) TransUNet; (e) DCUNet; (f) Ours.}
\label{chap4:fig:usm_seg}
\end{figure*}

The visual results of different methods on THU-MVFV3V dataset and USM dataset are shown in Fig. \ref{chap4:fig:thumvfv3v_seg} and Fig. \ref{chap4:fig:usm_seg} respectively. 
The vein masks from the THU-MVFV dataset typically appear thin and dispersed due to the selection criteria that only pixels with high reliability, as determined by MVA, are considered valid. In contrast, the vein masks in the USM dataset are characterized by their thickness and continuity and this feature is largely attributable to the human-annotated masks used during EIFNet's training.

Fig. \ref{chap4:fig:thumvfv3v_seg} demonstrates that while all listed methods proficiently segment long, continuous and thick veins presented in the THU-MVFV3V dataset, their performance varies significantly with thin and short veins. In Row No. 1, our method particularly excels at discerning the complex vein structures in the low-contrast region, as highlighted by the red box. In Row No. 2, it achieves the best performance in bridging the thin and fragile sections. 

\begin{table*}[htbp]
\centering
\caption{Comparison of Segmentation Results of Different Methods on USM and THU-MVFV3V Dataset}
\renewcommand\arraystretch{1.5}
\setlength{\tabcolsep}{0.0615\columnwidth}
\begin{tabular}{c|l|ccccc}
\hline
Dataset & Method & Dice(\%) & clDice(\%) & ACC(\%) & AUC(\%) & mIoU(\%) \\ \hline
\multirow{4}{*}{USM} 
& UNet \cite{cciccek20163d} & 77.75 & 82.73 & 83.41 & 84.32 & 63.60 \\
& TransUNet \cite{chen2021transunet} & 82.04 & 86.49 & 86.64 & 87.49 & 69.55 \\
& DCU-Net \cite{yang2022dcu} & 81.56 & 86.60 & 85.89 & 87.63 & 68.86 \\
& \textbf{Ours} & \textbf{84.35} & \textbf{93.30} & \textbf{88.18} & \textbf{89.33} & \textbf{72.94} \\ \hline
\multirow{4}{*}{THU-MVFV3V} 
& UNet \cite{cciccek20163d} & 77.96 & 85.14 & 84.98 & 94.24 & 63.88 \\
& TransUNet \cite{chen2021transunet} & 81.70 & 89.69 & 88.20 & 95.02 & 69.07 \\
& DCU-Net \cite{yang2022dcu} & 84.06 & 91.05 & 88.99 & 95.74 & 72.50 \\
& \textbf{Ours} & \textbf{86.47} & \textbf{93.16} & \textbf{92.22} & \textbf{96.16} & \textbf{76.17} \\ \hline
\end{tabular}
\label{chap4:tbl:all_seg}
\end{table*}

\begin{table*}[htbp]
\centering
\caption{Ablation Study on THU-MVFV3V Dataset}
\renewcommand\arraystretch{1.5}
\setlength{\tabcolsep}{0.04\columnwidth}
\begin{tabular}{c|cccc|ccc}
\hline
Combinations & Diffusion & Mask condition & FourierSSIM Loss & SD-Former & clDice(\%) & EER(\%) & ACC(\%) \\ \hline
Basic & \XSolidBrush & \XSolidBrush & \XSolidBrush & \XSolidBrush & 87.23 & 1.21 & 98.23 \\
$C_1$ & $\checkmark$ & \XSolidBrush & \XSolidBrush & \XSolidBrush & 87.57 & 1.08 & 98.50   \\
$C_2$ & $\checkmark$ & $\checkmark$ & \XSolidBrush & \XSolidBrush & 88.12 & 0.86 & 98.94   \\
$C_3$ & $\checkmark$ & $\checkmark$ & $\checkmark$ & \XSolidBrush & 90.02 & 0.62 & 99.69   \\
$C_4$ & $\checkmark$ & $\checkmark$ & \XSolidBrush & $\checkmark$ & 91.51 & 0.45 & 99.42   \\
Full & $\checkmark$ & $\checkmark$ & $\checkmark$ & $\checkmark$ & 93.16 & 0.33 & 99.71   \\ \hline
\end{tabular}
\label{chap4:fig:ablation}
\end{table*}

Fig. \ref{chap4:fig:usm_seg} demonstrates that while all predicted vein masks capture the overall topology of the finger veins in the USM dataset, they differ in how they represent the interconnections between vein streams. 
In Row No. 1, the ground truth mask shows that the veins within the red box should form a ring-like structure; our method is the only one that accurately depicts this feature. 
In Row No. 2, the vein originating from the left corner of the red box is expected to split into two distinct pathways. Other methods struggle to replicate this bifurcation, whereas our method's predicted mask exhibits the clearest visual representation of this detail. 

In conclusion, while the masks produced by various methods account for geometric aspects of vein paths, such as width and direction, our DiffVein stands out by generating masks with superior continuity. 
It achieves the best visual effect by predicting potential venous connections and replicating unique topological structures.
\subsection{Ablation Study \label{chap4:sec:ablation}}

In this section, we carried out extensive ablation experiments to evaluate the main modules we have proposed in this paper. As indicated in the Table \ref{chap4:fig:ablation}, a factorial experiment including six combinations of factors is designed:

\begin{itemize}
\item \textit{\textbf{Basic}} is the baseline which removes the modules related to diffusion and keeps only the segmentation branch.
\item \textit{\textbf{C$_1$}} only applies the vanilla diffusion process onto the segmentation branch without introducing all the other modules.
\item \textit{\textbf{C$_2$-C$_4$}} includes the specific modules as ticked in the table. When FourierSIM loss is not ticked, merely MSE loss is used for noise estimation, otherwise both MSE and FourierSIM losses are applied.
\item \textit{\textbf{Full}} includes all the modules we discussed as mentioned above in this paper.
\end{itemize}

Table \ref{chap4:fig:ablation} reports the clDice, EER and ACC of DiffVein and the other five ablation configurations on the THU-MVFV3V dataset. 

\subsubsection{Effect of diffusion model}
Compared to \textit{\textbf{Basic}}, \textit{\textbf{C$_1$}} demonstrates an improvement in EER by 0.13\% and in ACC by 0.27\%, indicating the diffusion model can improve the overall performance of finger vein authentication task by learning better feature representations. However, no significant improvement is observed in the clDice score. This may be attributed to the fact that the segmentation network learns exclusively from the forward process without receiving any feedback from the denoising branch.

\subsubsection{Effect of mask condition}
\textit{\textbf{C$_2$}} outperforms \textit{\textbf{C$_1$}} in the finger vein authentication task, as evidenced by a 0.22\% reduction in EER and a 0.44\% increase in ACC. This suggest that incorporating the mask condition into the denoising process provides valuable supervisory information. Furthermore, it demonstrates that the feature map associated with the vein mask can act as an effective conditional guidance in the denoising network. However, in segmentation tasks, the absence of an information feedback mechanism in this set of experiments resulted in only a modest improvement in the clDice score.

\subsubsection{Effect of FourierSIM loss}
\textit{\textbf{C$_3$}} witnessed a notable improvement in EER by 0.24\% compared to \textit{\textbf{C$_2$}}, suggesting that the FourierSIM loss significantly enhances the denoising network's performance within the diffusion model. FourierSIM loss extends the supervised learning process in Euclidean-space to Fourier-space, thereby providing additional spectrum information. Consequently, the encoder within the segmentation network also produces better feature representations, inadvertently leading to improved segmentation performance. Another separate set of comparative experiments between \textit{\textbf{C$_4$}} and \textit{\textbf{Full}} also supports the conclusion mentioned above.

\subsubsection{Effect of SD-Former}
\textit{\textbf{Full}} yields a 0.29\% reduction in EER and a 0.97\% increase in ACC over \textit{\textbf{C$_3$}}, along with a 3.14\% rise in the clDice score. These improvements underscore the benefit of leveraging embeddings from the denoising process to augment the segmentation task. The introduction of the SD-Former allows the segmentation network to assess and refine the feature vectors generated by its encoder using feedback from the denoising branch. This bidirectional exchange of information develops a cooperative learning manner. Our findings confirm that the mutual interaction between the segmentation and denoising networks contributes to superior performance in both tasks examined in this study.
\section{Conclusion}
In this work, we introduced a novel diffusion network called DiffVein which tackles both finger vein segmentation and authentication tasks concurrently. To the best of our knowledge, this is the first application of a generative diffusion model in the field of finger vein recognition. DiffVein is composed of a segmentation branch for vein mask generation, and a diffusion branch which focuses on finger vein classification through a series of forward diffusion and denoising steps. Within the segmentation branch, we presented a novel feature extraction module termed mask condition to offer vein semantic information as conditional guidance to the denoising network. Additionally, we proposed the Semantic Differential Transformer (SD-Former), a feature fusion module designed to extract category semantics-related features within the denoising network. The SD-Former employs a Fourier-space attention mechanism to effectively separate different semantics information from various spectral components. Furthermore, we proposed a novel loss function named Fourier-space Structural Similarity (FourierSIM) loss. This function supervises the denoising network by evaluating the structural similarity of noise characteristics, such as amplitude and phase, in the Fourier domain. Our extensive comparative experiments show DiffVein's superior performance in both segmentation and authentication tasks against other methods. The results of ablation studies validate the significant contribution of our proposed modules to DiffVein's efficacy, facilitating a mutual improvement in finger vein segmentation and authentication.


%

\ifCLASSOPTIONcaptionsoff
  \newpage
\fi



%
\bibliographystyle{IEEEtran}
\bibliography{main}

%





\end{document}